\documentclass[sigplan,10pt]{acmart}

\usepackage{amsmath}
\usepackage{xspace}
\usepackage{caption}
\usepackage{subcaption}
\usepackage{float}
\usepackage{pifont}
\usepackage{algorithm}
\usepackage{algpseudocode}
\usepackage{amsmath}
\algrenewcommand\algorithmiccomment[1]{\hfill$\triangleright$~#1}

\newcommand{\heading}[1]{\vspace{4pt}\noindent\textbf{#1}}

\newcommand{\ie}{i.e.\xspace}
\newcommand{\sysname}{Akashic\xspace}

\renewcommand\footnotetextcopyrightpermission[1]{}

\makeatletter
\def\@copyrightspace{}
\makeatother

\renewcommand{\shortauthors}{}

\setcopyright{none}
 \author{
 {\rm Yang Liu}\textsuperscript{$\ast$}$^\dagger$ \quad
 {\rm Zhaokai Luo}\textsuperscript{$\ast$}$^\dagger$\quad
 {\rm Huayi Jin}$^\dagger$ \quad
 {\rm Ruozhou He}$^\dagger$ \quad
 {\rm Chenchen Hong}$^\dagger$ \quad
 {\rm Zhiyong Wang}$^\dagger$ \quad\\
 {\rm Yifei Liu}$^\mathparagraph$ \quad
 {\rm Yunfei Gu}$^\mathparagraph$ \quad
 {\rm Chentao Wu}$^\mathparagraph$  \quad
 {\rm Junhao Hu}$^\mathsection$\\
 $^\dagger${\it Xiaohongshu Inc., China} \quad
 $^\mathparagraph${\it ShangHai JiaoTong University}\quad 
 $^\mathsection${\it Peking University}\\[0.6ex]
 {\normalsize\rm
   \textsuperscript{$\ast$}Corresponding to:\;
   Zhaokai Luo~\href{mailto:luozhaokai@xiaohongshu.com}{\textless luozhaokai@xiaohongshu.com\textgreater}%
 }\\
}
\begin{document}
\begin{abstract}
Recent LLM-based agent systems continuously accumulate context across multi-turn interactions, tool invocations, and cross-session workflows.   ~Replaying the full history for every request quickly becomes impractical: long contexts increase prefill cost, may exceed context limits, and often bury task-relevant evidence in irrelevant content, degrading both serving efficiency and output quality.
We propose \sysname, a low-overhead memory system built around MemAttention, which organizes context into bounded chunks and models semantic relationships across chunks, preserving cross-chunk evidence without repeatedly rewriting the full history. \sysname further applies hardware--software co-designed memory placement to co-locate likely co-retrieved chunks, reducing retrieval fragmentation and I/O overhead. Across four representative workloads and three model sizes, \sysname improves task accuracy by up to 10.2 points, throughput by up to 1.21$\times$, and sustainable request rate by up to 1.88$\times$ over strong prior memory baselines.
\end{abstract}
\title{Akashic: A Low-Overhead LLM Inference Service with MemAttention}
\maketitle
\thispagestyle{plain}
\pagestyle{plain}

\section{Introduction}
\label{sec-intro}
\begin{figure}[t]
    \centering
    \includegraphics[width=.98\columnwidth]{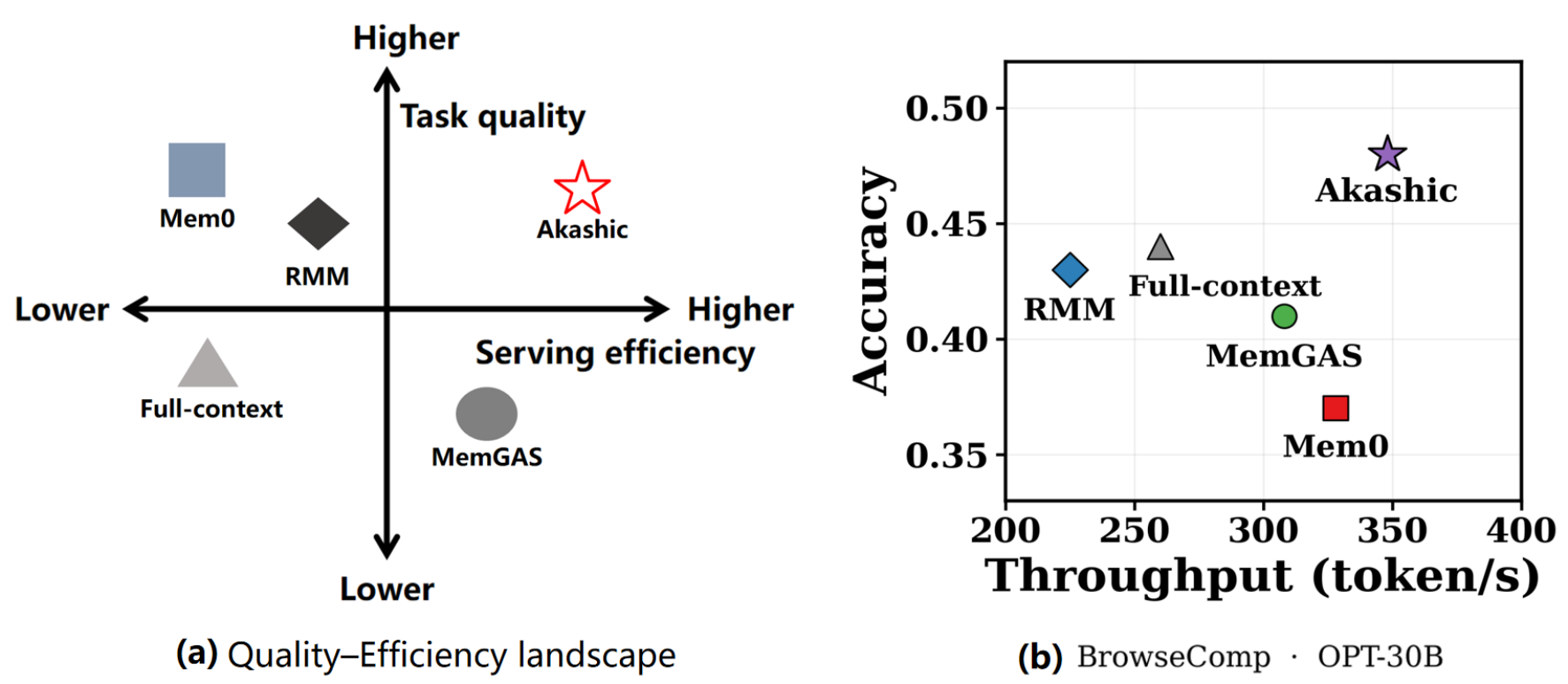}
    \caption{(a)~Existing memory designs occupy different points on the task-quality / serving-efficiency trade-off, whereas \sysname targets the high-quality, high-efficiency regime. (b)~On BrowseComp with OPT-30B, \sysname improves both accuracy and throughput over all baselines, outperforming the strongest prior method by about 2.0 points in accuracy and about 1.35$\times$ in throughput.}
    \label{fig:introduce}
    \vspace{-20pt}
\end{figure}

Recent agent systems continuously accumulate input context across multi-turn interactions, tool invocations, and cross-session execution, and this growing context can substantially degrade model throughput and output quality~\cite{fan2023llmse,chen2021evaluating,copilot,maharana2024locomo,packer2023memgpt,cobbe2021gsm8k,wei2022chainofthought}. A straightforward solution is to feed the full interaction history into every request, but this approach quickly becomes impractical: long prompts increase prefill cost, can exceed context limits, and often bury task-relevant evidence in irrelevant context, hurting both serving efficiency and output quality~\cite{liu2024lostinthemiddle,shaham2023zeroscrolls}. As a result, modern agent systems increasingly maintain external memory and re-inject only a subset of prior context at inference time~\cite{packer2023memgpt,mem0,anthropic_claude_code}.

A common way to build such memory is \emph{summarization}: once the accumulated history exceeds a threshold, the system compresses older context into a shorter representation and uses that representation in future requests~\cite{langchain_conversationsummarymemory,llamaindex_memory_doc}. Existing summarization-based designs, however, expose a granularity trade-off. \emph{Whole-context} approaches such as Mem0 summarize the entire history at each trigger point~\cite{mem0}. They are simple, but their update cost grows with history length, and unrelated topics are mixed into one memory object, which can inject noise at retrieval time. \emph{Segment-level} approaches such as MemGAS summarize smaller units independently~\cite{xu2025memgas}. They bound per-update cost, but semantically dependent evidence may be split across segments and never recovered together, weakening multi-turn and long-range reasoning.

In practical LLM serving, resolving this trade-off requires addressing two key challenges. First, context information density is highly non-uniform across workloads and even across different phases of the same interaction~(\S\ref{sec:3.1}): some spans are highly compressible, whereas others are dense and expensive to compress. A fixed global compression policy is therefore often mismatched to the actual workload. Second, agent memory is not only a semantic object but also a storage object. Even when retrieval selects only a few relevant memories, end-to-end latency can remain high if those memories are physically scattered across many pages or blocks. We refer to this mismatch between semantic relatedness and physical placement as the \emph{locality gap}~(\S\ref{sec:locality-gap-agent-memory}).

We present \textbf{\sysname}, a low-overhead memory system for LLM serving that addresses these challenges through \textbf{MemAttention} and \textbf{hardware--software co-designed} memory placement optimization. \sysname performs \emph{chunk-granular} memory maintenance: it compacts one bounded chunk at a time, rather than repeatedly rewriting the full history. To preserve cross-chunk evidence, \sysname uses \emph{cross-chunk inference} to reconcile each new chunk with a small set of semantically related prior chunks before writing the result back to memory. This design bounds maintenance overhead while retaining information that would otherwise be fragmented by independent segment summaries~(\S\ref{sec:MemAttention}). \sysname further includes a \emph{hardware--software co-optimized memory manager} that co-locates chunks likely to be co-retrieved and compacts stale data out of place, reducing read amplification and retrieval-side contention during concurrent serving~(\S\ref{sec:block-space-manager}).

We evaluate \sysname on four representative benchmarks spanning diverse workload characteristics---LoCoMo~\cite{maharana2024locomo}, SWE-bench~\cite{jimenez2024swebench}, BrowseComp~\cite{wei2025browsecomp}, and WebArena~\cite{zhou2024webarena}---using models of different sizes. As shown in Figure~\ref{fig:introduce}(a), \sysname consistently achieves both higher serving efficiency and better task quality than prior baselines. Figure~\ref{fig:introduce}(b) further illustrates this advantage with a representative example, where \sysname substantially outperforms existing methods in both accuracy and throughput~(\S\ref{sec:eval}). \sysname consistently lies on the \emph{Pareto frontier}, improving task accuracy by up to 10.2 points and throughput by up to 1.21$\times$ over the strongest memory baselines under the same experimental setting, while sustaining up to 1.88$\times$ higher request rates under concurrent serving.We will continue to update Akashic on newer models and datasets.  
The subsequent code will be merged into an alternative branch of the RedKnot framework~\cite{redknot}.  
The open-source address is \url{https://github.com/rednote-machine-learning/RedKnot}.
\begin{itemize}
    \item We characterize two key bottlenecks in agent memory: the mismatch between fixed summarization policies and heterogeneous context density, and the locality gap between semantic relevance and physical placement.
    \item We design \emph{MemAttention}, a chunk-granular memory maintenance mechanism that reconciles each new chunk with a small set of semantically related prior chunks, preserving cross-chunk evidence while bounding maintenance overhead.
    \item We propose and implement a \emph{hardware--software co-designed memory manager} to mitigate memory fragmentation in LLM serving, by co-locating likely co-retrieved chunks and compacting stale data to reduce retrieval I/O and contention.
    \item We implement \sysname and evaluate it on four representative workloads, showing up to 10.2-point higher accuracy, 1.21$\times$ higher throughput, and 1.88$\times$ higher sustainable request rate than the strongest prior memory baselines.
\end{itemize}

\section{Background}
\label{sec:background}

\subsection{Long Context as a First-Class Bottleneck}
\label{sec:bg-long-context}

Modern agentic applications (e.g., multi-round assistants, tool-augmented planners, and long-running workflows) naturally accumulate conversational state over time, pushing the input context toward the model's maximum context window. Although recent frontier models have dramatically expanded their nominal context limits (up to million-token regimes), long-context requests remain expensive in latency and compute: for example, OpenAI reports that time-to-first-token (TTFT) can increase from seconds at $\sim$128K tokens to around a minute at $\sim$1M tokens, even with an optimized inference stack~\cite{openai_gpt41,openai_gpt41_modeldoc}.\\
\indent Beyond the hard limit of the maximum window, longer context also degrades \emph{effective task accuracy}. Empirically, LLMs exhibit strong attention biases when relevant evidence appears in the middle of long prompts: performance can follow a distinctive U-shaped curve and drop sharply when the answer-bearing content is placed mid-context~\cite{liu2024lostinthemiddle}. Subsequent work shows that even with long-context model variants, robust middle-of-context utilization remains challenging, motivating positional encoding and long-context evaluation efforts~\cite{zhang2024mspoe,shaham2023zeroscrolls}. For long-term dialogue specifically, benchmarks such as LoCoMo demonstrate that very long multi-session conversations amplify temporal and causal reasoning failures, and that long-context or retrieval-augmented approaches still lag substantially behind human-level consistency~\cite{maharana2024locomo}.\\
\indent During serving of LLM inference, long-context requests are expensive primarily because the Transformer must execute a full \emph{prefill} pass over the entire prompt to construct per-layer activations and populate the KV cache, incurring \textbf{substantial compute and memory traffic}. At the systems layer, this prefill cost grows rapidly with sequence length: standard Transformer attention exhibits quadratic compute and memory scaling in the prompt length~\cite{vaswani2017attention}, and even exact attention implementations require IO-aware kernel designs to prevent memory bandwidth from becoming the dominant performance limiter~\cite{dao2022flashattention}. These constraints interact strongly with agent memory.
\subsection{Memory Compression and Retrieval in Practice}
\label{sec:back memory}
To cope with bounded windows and long-context degradation, common agent frameworks (such as Mem0~\cite{mem0}, RMM~\cite{rmm}, and SeCom~\cite{pan2025secom}) adopt \emph{compression} (summarization) and \emph{selective retrieval}. A representative pattern maintains a \emph{running summary} that is updated incrementally after each turn (or when a token threshold is reached), and injects only the summary (optionally plus a short recent buffer) into future prompts; this pattern is explicitly implemented by widely used framework abstractions. Complementarily, retrieval-based memory stores past messages (or derived facts) in a vector index and retrieves a small set of semantically relevant items to condition the model, which reduces prompt length but introduces retrieval and consolidation challenges. This end-to-end pipeline of memory compression, vector-database storage, and inference-time retrieval constitutes the \emph{memory workflow} in modern LLM agents.

Recent open-source systems elevate memory to an OS-inspired hierarchy. MemGPT formulates \emph{virtual context management} and pages information between in-context working memory and external storage, enabling long-running agents beyond fixed context limits~\cite{packer2023memgpt}. Industrial and open-source memory layers such as Mem0 similarly emphasize compressing history into compact memory representations and retrieving only a few relevant memory items for each user query~\cite{mem0_github}. Tooling ecosystems such as LlamaIndex increasingly expose configurable memory modules that combine a FIFO short-term buffer with customizable long-term memory blocks for extraction and reinjection~\cite{llamaindex_memory_doc}. Overall, existing practice converges on a trade-off: frequent summarization improves boundedness but consumes additional model calls, whereas infrequent summarization risks window overflow and accuracy loss---a tension that motivates more controllable, fine-grained, and update-friendly memory organization.

\section{Memory Challenges in LLM Serving}
\label{sec:motivation}
\begin{figure}[t]
    \centering
\includegraphics[width=0.98\columnwidth,height=0.75\textheight,keepaspectratio]{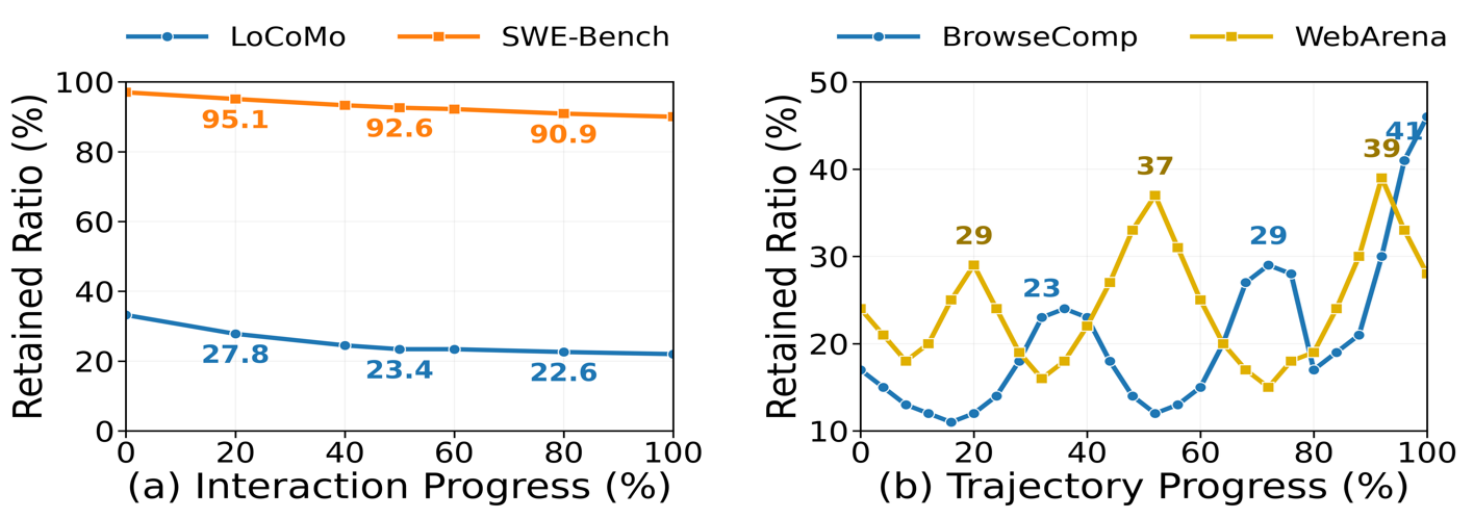}
\caption{Context information density varies substantially across datasets and over time. (a) Compared with LoCoMo, SWE-Bench exhibits a higher retention ratio. (b) BrowseComp and WebArena show bursty and alternating retention ratio within trajectories.}
 \vspace{-10pt}
\label{fig:Retained Ratio}
\end{figure}
\subsection{Memory Inefficiency under Heterogeneous Context Density}
\label{sec:3.1}
\heading{Non-Uniform Context Information Density.}
Extracting effective information from a long input context is itself an inference process. Before an LLM system can benefit from a compressed memory representation, it must first identify which parts of the raw context are salient, and this step consumes additional computation. Let $L_{\mathrm{raw}}$ denote the token length of the original context and $L_{\mathrm{eff}}$ denote the token length of the effective information preserved after memory construction or compression. We define the \emph{Retained Ratio} as
\begin{equation}
\rho = \frac{L_{\mathrm{eff}}}{L_{\mathrm{raw}}}.
\end{equation}
A smaller $\rho$ indicates that the context is highly compressible, whereas a larger $\rho$ indicates that most tokens are information-bearing and therefore difficult to compress.

As shown in Figure~\ref{fig:Retained Ratio}, context information density is highly non-uniform, both across datasets and over time within the same workload. Figure~\ref{fig:Retained Ratio}(a) shows a clear cross-dataset contrast: LoCoMo has a much lower retained ratio than SWE-Bench, indicating that its long conversational histories contain substantial redundancy and are therefore more amenable to traditional compression-based memory strategies. In contrast, SWE-Bench remains consistently high in retained ratio, suggesting that most tokens are information-bearing, as is common in code debugging and repository-level reasoning. In such high-density workloads, compression still triggers an additional inference step for summarization or memory construction, but yields only limited reduction in prompt length. As a result, memory maintenance may introduce extra overhead without meaningfully shortening the context, reducing or even negating the end-to-end system benefit. Figure~\ref{fig:Retained Ratio}(b) further shows that information density is often unstable even within a single trajectory. BrowseComp and WebArena exhibit bursty and alternating retained ratios, with some phases being highly compressible and others being highly information-dense. This temporal heterogeneity makes static memory policies fundamentally mismatched to agent workloads. A fixed summarization trigger, compression ratio, or retrieval budget may be beneficial during low-density phases but wasteful during high-density phases. Consequently, conventional memory workflows struggle to deliver stable gains across the full lifetime of an interaction.
\begin{figure}[t]
    \centering
\includegraphics[width=0.98\columnwidth,height=0.75\textheight,keepaspectratio]{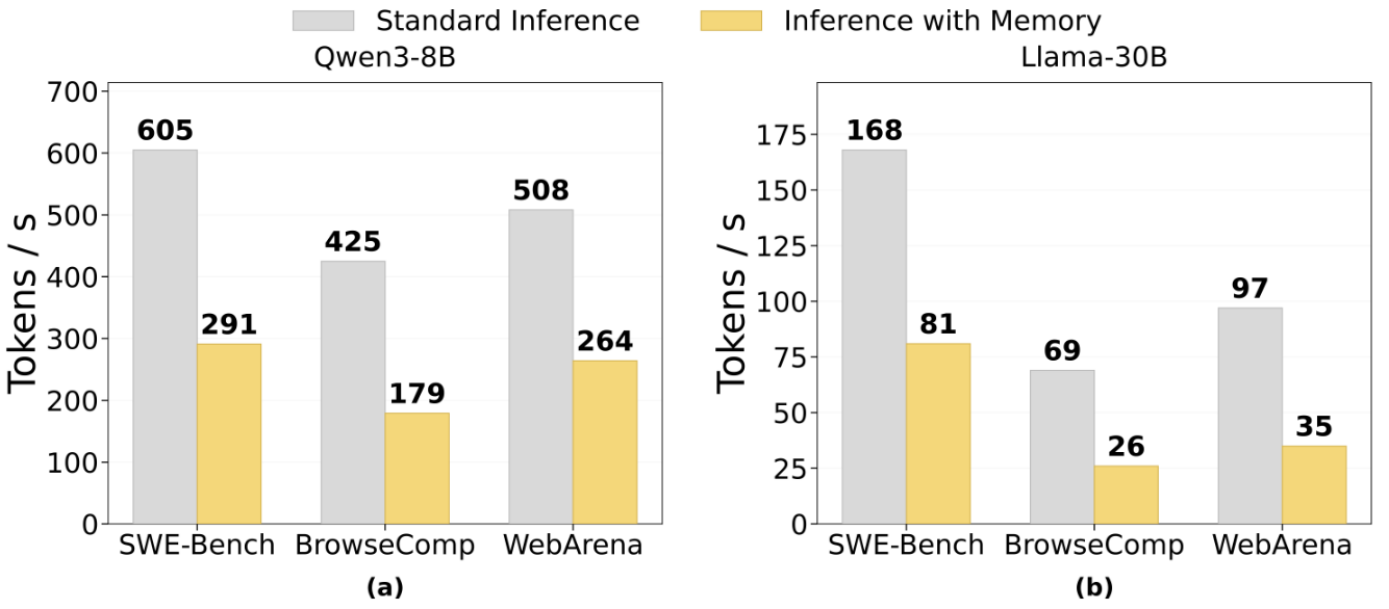}
\caption{Memory-augmented inference reduces end-to-end throughput by 48.0\%--63.9\% across SWE-Bench, BrowseComp, and WebArena.}
 \vspace{-10pt}
\label{fig:Memory-augmented inference}
\end{figure}
This observation has direct systems implications. Representative long-term memory methods---including recursive summarization~\cite{recursive}, MemGPT~\cite{packer2023memgpt}, SeCom~\cite{pan2025secom}, Reflective Memory Management (RMM)~\cite{rmm}, and Mem0~\cite{mem0}---maintain usable memory representations through summary generation, memory extraction, hierarchical memory management, segmentation, denoising, or reflective consolidation. These operations are beneficial when $\rho$ is low, because a relatively short memory representation can replace a much longer raw context. However, when $\rho$ is high, the benefit of compression becomes marginal. For example, compressing a 1000-token context into 900 tokens yields only a small reduction in prompt length, yet still incurs the full cost of an additional memory-maintenance step. In such high-density regimes, memory construction may save little context while introducing extra inference overhead, ultimately wasting compute and reducing end-to-end serving throughput.\\
\heading{Inference Resource Waste under High Context Density.} As discussed in \S\ref{sec:back memory}, the memory workflow first feeds the context into the LLM for inference. The resulting memory is then stored in a database, and relevant memory context is retrieved and loaded during the next actual inference. We evaluate SWE-Bench, BrowseComp, and WebArena, which represent workloads with either high information density or highly unstable information-density patterns, on an NVIDIA H20 96GB GPU using SGLang 0.4.6.post1~\cite{sglang_046post1} with batch size 1.
As shown in Figure~\ref{fig:Memory-augmented inference}, end-to-end throughput decreases substantially on datasets with unstable, low information density. The reason is twofold: memory generation requires extra computation during inference, and memory management further incurs database write and retrieval overheads, both of which prolong the inference pipeline. Because context compressibility is not directly observable beforehand and often depends on downstream queries, accurately deciding when and how aggressively to compress requires explicit modeling of context informativeness and query relevance~\cite{li-etal-2023-compressing,jiang-etal-2023-llmlingua,an-etal-2025-lcirc}; such an adaptive policy is therefore difficult to realize in practice. Instead of attempting global compression over the entire context, we adopt chunk-granular compression. By compressing only one chunk at a time, we bound the per-step inference overhead and make the cost of memory construction more stable and controllable.
\subsection{The Locality Gap in Agent Memory}
\label{sec:locality-gap-agent-memory}
\begin{figure}[t]
    \centering
\includegraphics[width=0.98\columnwidth,height=0.75\textheight,keepaspectratio]{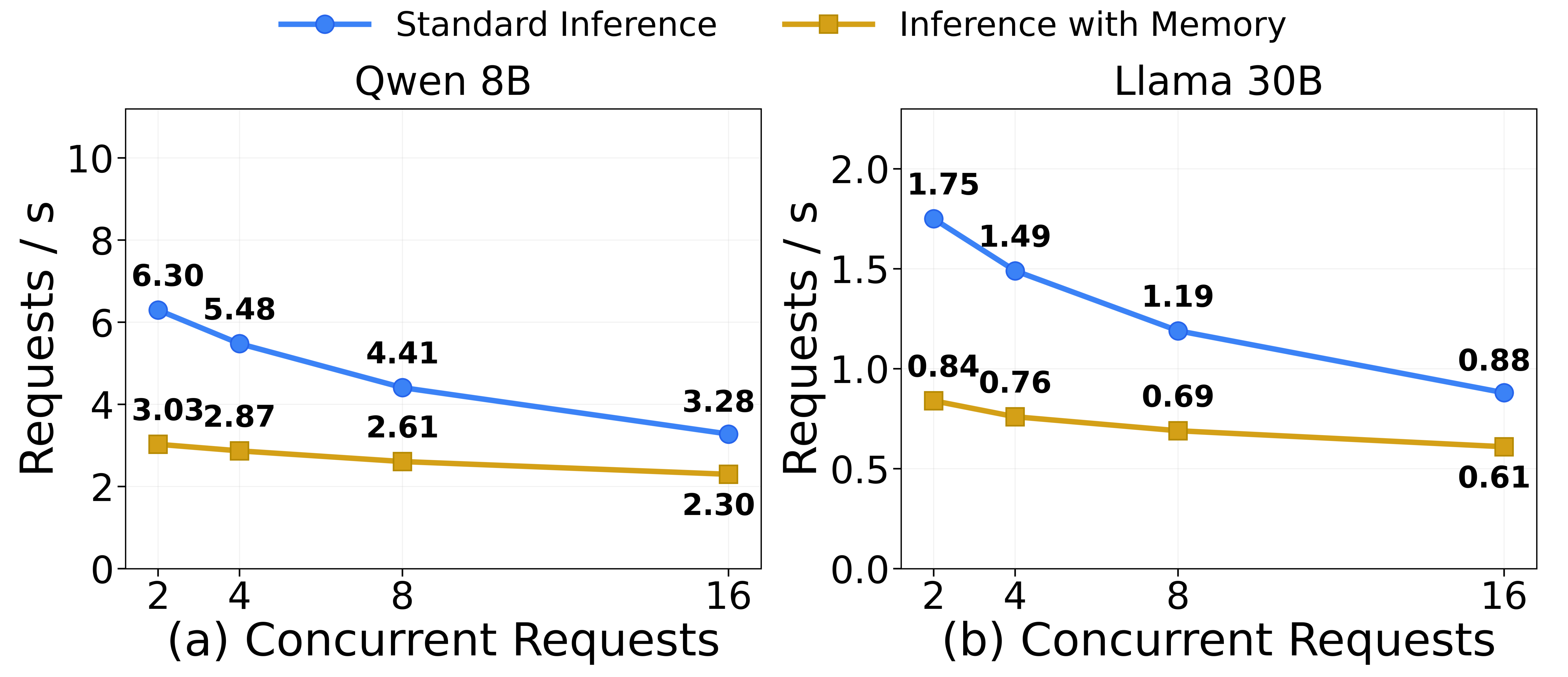}
\caption{Throughput on LoCoMo as concurrency increases. Although memory-augmented inference consistently underperforms standard inference, its throughput gradually approaches the no-memory baseline at larger batch sizes.}
\vspace{-10pt}
\label{fig:Locality Gap}
\end{figure}
Existing agent-memory systems are primarily designed around a semantic objective: deciding what to retain, how to structure it, and which memories to retrieve at inference time. MemGPT introduces a tiered memory hierarchy for long-horizon interaction~\cite{packer2023memgpt}; RMM improves long-term dialogue memory through multi-granular summarization and retrieval refinement~\cite{recursive}; and A-MEM dynamically links and evolves memories through agentic indexing~\cite{xu2025amem}. These systems substantially improve memory quality, but they largely treat storage as an opaque backend. As a result, semantic relatedness and physical placement are optimized separately.

This separation creates what we call the \emph{locality gap}. Let $R_t$ denote the set of memories retrieved at time $t$, and let $P(R_t)$ denote the set of storage pages or blocks touched to materialize $R_t$. Existing memory managers primarily optimize the semantic quality of $R_t$, whereas end-to-end retrieval cost is dominated by $|P(R_t)|$. The locality gap is therefore large when $|R_t|$ is small but $|P(R_t)|$ remains large. This is precisely the failure mode identified in disk-resident vector search. SPANN explicitly reduces the number of disk accesses in hybrid memory--disk ANN search~\cite{chen2021spann}; Starling improves performance by reordering the disk-resident graph layout to enhance locality and reduce bandwidth waste~\cite{wang2024starling}; and recent work on SSD-backed vector databases shows that frequently co-accessed nodes are often placed on different pages, making locality-preserving colocation critical for reducing I/O overhead~\cite{shim2025turbocharging}. The implication for agent memory is direct: retrieving only a few relevant memories does not imply low latency if those memories are physically scattered.

To quantify this effect, we use the same experimental environment as in ~\S\ref{sec:3.1}. We evaluate LoCoMo, a long-horizon workload in which semantically related evidence is often distributed across distant interaction steps, while scaling concurrency from 2 to 16 requests. As shown in Figure~\ref{fig:Locality Gap}, memory-augmented inference consistently remains below standard inference on both Qwen-8B and Llama-30B, and the throughput gap gradually narrows as concurrency increases. On Qwen-8B, the throughput ratio between standard inference and inference with memory drops from 2.08$\times$ at 2 concurrent requests to 1.43$\times$ at 16 concurrent requests. On Llama-30B, the same ratio drops from 2.08$\times$ to 1.44$\times$. Equivalently, memory-augmented inference rises only from 48.1\% to 70.1\% of the no-memory baseline on Qwen-8B, and from 48.0\% to 69.3\% on Llama-30B. This trend is consistent with worsening locality under concurrent serving: although the retrieved memory set is semantically selective, the corresponding memory items can still be physically scattered across many pages or blocks. As concurrency grows, the system must materialize more such scattered accesses at the same time, increasing retrieval-side contention and progressively offsetting the semantic benefit of shorter prompts. In other words, semantic selectivity alone is insufficient when co-retrieved memories are not placed with good physical locality.

The problem is amplified by the write pattern of agent memory. Unlike static indexes, agent memories are continuously appended, summarized, linked, and updated over long-lived interactions~\cite{recursive,xu2025amem}. Under such dynamic writes, any locality that happens to exist today can quickly deteriorate tomorrow. More broadly, the ANN literature shows that maintaining freshness under real-time updates is itself nontrivial, because search efficiency and update cost must be balanced rather than optimized in isolation~\cite{singh2021freshdiskann}. Therefore, optimizing only semantic relevance is insufficient for scalable agent memory. What is needed is an affinity-aware memory manager that jointly reasons about \emph{which} memories are likely to be co-retrieved and \emph{where} they should be placed, so that future retrievals touch fewer pages while rewrite, compaction, and maintenance overhead remain bounded.
\section{Design of \sysname}

\begin{figure*}[t]
    \centering
\includegraphics[width=0.9\textwidth,height=0.4\textwidth]{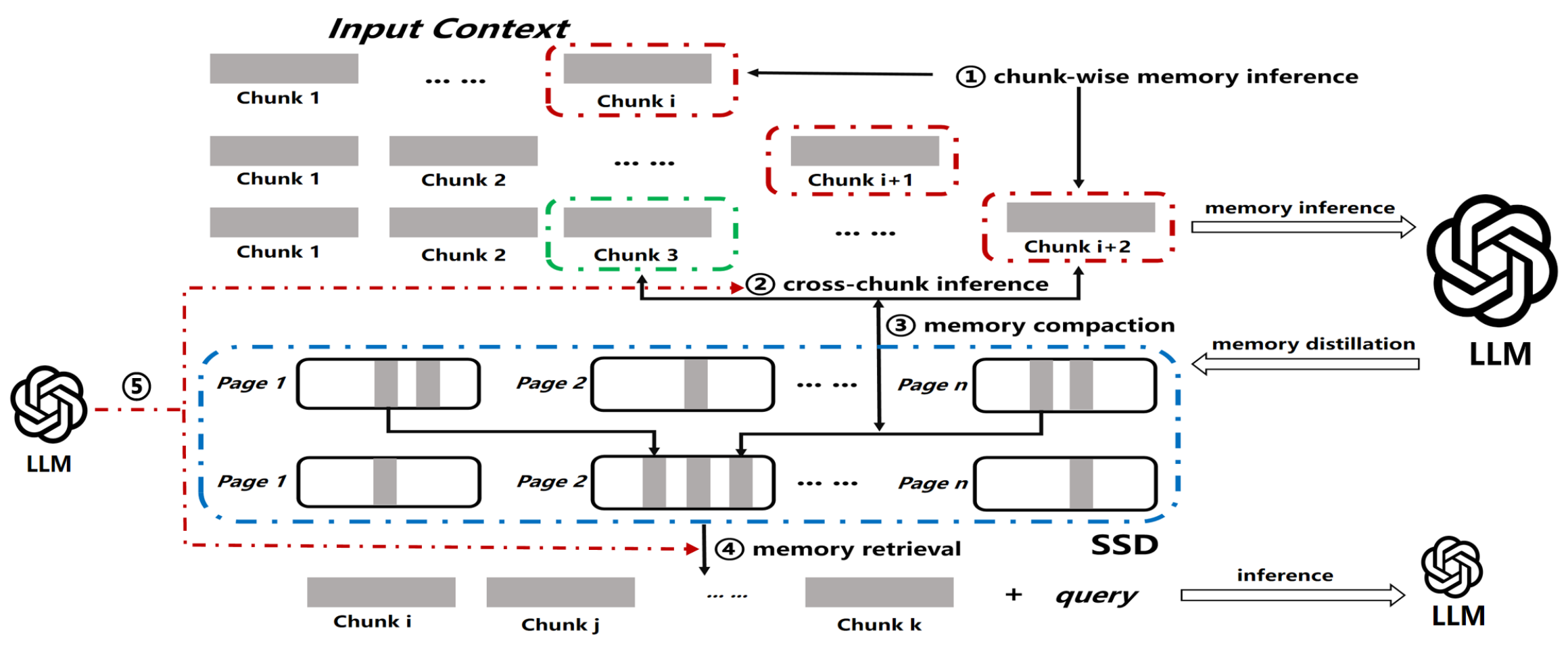}
    \vspace{-4pt}
    \caption{\sysname system overview.}
    \label{fig:system-overview}
    \vspace{-8pt}
\end{figure*}

In this work, we develop a new attention algorithm for memory, \emph{MemAttention}, and build an LLM serving engine, \emph{\sysname}, to tackle the challenges outlined in \S\ref{sec:motivation}. The architecture of \sysname is shown in Figure~\ref{fig:system-overview}.
\ding{182}~\sysname adopts chunk-granular memory summarization, treating each chunk as the basic unit of context. \ding{183}~For every new chunk, it retrieves previously seen chunks with relevant information and performs joint summarization over the associated chunks. \ding{184}~Chunk compaction organizes related memories into contiguous physical locations and removes irrelevant ones. \ding{185}~Memory retrieval selects the memory chunks relevant to the current inference and incorporates them into the input context. \ding{186}~The model identifies the relatedness of memory chunks.\\
\indent Next, we describe the MemAttention algorithm in \S\ref{sec:MemAttention}. We then present the design of the memory-chunk manager in \S\ref{sec:block-space-manager} and show how it supports MemAttention. Finally, we describe the implementation of \sysname in \S\ref{sec:impl}.
\subsection{MemAttention}
\label{sec:MemAttention}
\begin{figure}
    \centering
    \includegraphics[width=.95\columnwidth]{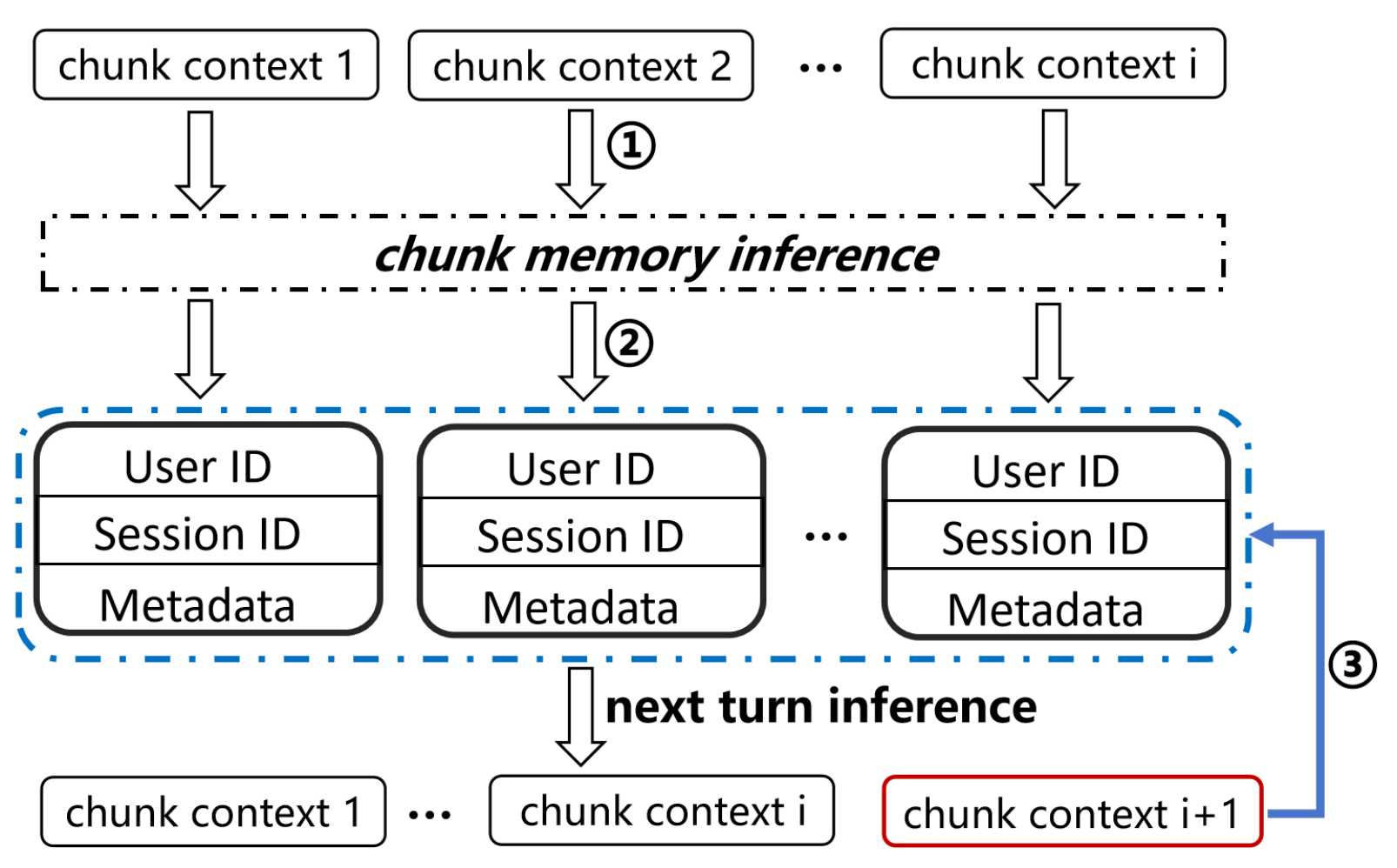}
    \caption{Workflow of MemAttention: each incoming chunk context is incrementally compacted into a structured memory record with user- and session-scoped metadata, and the most relevant prior chunk memories are retrieved to augment the next-turn inference. }
    \label{fig:MemAttention}
    \vspace{-10pt}
\end{figure}

To address the memory challenges in \S\ref{sec:3.1}, we introduce \textbf{MemAttention}, a chunk-granular memory architecture that treats memory construction as incremental maintenance rather than global compression over the entire interaction history. MemAttention stores memory as chunk objects keyed by $(\texttt{user\_id}, \texttt{session\_id}, \texttt{metadata})$, where \texttt{user\_id} isolates tenants, \texttt{session\_id} preserves conversation boundaries, and \texttt{metadata} records representative information of a chunk within a session. For each active session, the system maintains one append-only active chunk for newly arrived turns and a set of compacted chunks for prior history. As shown in Figure~\ref{fig:MemAttention}, \ding{182}~we first partition the input context into fixed-size chunks. Unless otherwise specified, we empirically set the chunk size to 1024 tokens~(\S\ref{Sec:5.5}), which provides a good balance between memory quality and maintenance overhead.
Once the accumulated context reaches one chunk, the system performs \emph{chunk memory inference} over that chunk in order to extract its salient information.
\ding{183}~The inferred result is materialized as a \emph{memory chunk} object indexed by \texttt{user\_id}, \texttt{session\_id}, and metadata, which respectively preserve tenant isolation, session boundaries, and temporal order.
\ding{184}~At the next turn, newly arrived context is accumulated and partitioned with the same chunk granularity, and each completed chunk is processed in the same way to generate a new memory chunk incrementally.
In this way, MemAttention treats memory construction as continuous chunk-level maintenance rather than global compression over the entire interaction history.

The key design principle is \textbf{gate-triggered chunk compaction}. Rather than reprocessing the full history whenever new context arrives, MemAttention triggers compaction only when the active chunk reaches a predefined threshold, i.e., when $L_{\text{chunk}} \ge \tau_c$, where $L_{\text{chunk}}$ is the token length of the active chunk and $\tau_c$ is the chunk gate. Once triggered, the model rewrites only that chunk into a compacted representation that co-locates semantically related memory items and removes transient or redundant content. This design does not require estimating context density \emph{a priori}. Whether the incoming context is information-dense or highly compressible, the worst-case maintenance cost of a single compaction step scales as $O(L_{\text{chunk}})$ with $L_{\text{chunk}} \le \tau_c$, rather than with the total history length. In this sense, MemAttention follows a common systems design pattern: maintenance is triggered by local thresholds and applied to bounded units, analogous to segment cleaning in log-structured file systems, rolling merges in LSM-trees, and size-triggered compactions in Bigtable~\cite{rosenblum1992lfs,oneil1996lsm,chang2006bigtable}.

However, compacting one memory chunk at a time makes it difficult to update or remove metadata associated with previously compacted chunks. For example, newly arrived information may overwrite or invalidate older memories, yet the stale metadata of earlier chunks may still persist, causing conflicting information when multiple memory chunks are loaded for inference. To address this issue, we introduce \emph{cross-chunk inference}, inspired by the efficient memory retrieval strategy used in Claude Code~\cite{claude_code_sourcemap_github}. Specifically, the system feeds the model with the keywords of the chunk to be compacted together with the metadata keywords of previously compacted chunks, and \emph{asks the model to perform semantic matching across them}, thereby identifying highly related memory chunks for joint reasoning. We then select the top-5 most relevant chunks, following Claude Code, for cross-chunk inference, which enables the system to update obsolete memories and remove invalidated ones. After memory chunks are compacted, MemAttention performs \emph{query-aware retrieval} during inference. For a new input context, the system first collects candidate memory chunks within the current \texttt{user\_id} and \texttt{session\_id} namespace and retrieves their metadata keywords. Rather than relying on predefined similarity scores, such as dense embedding similarity or token-overlap matching, MemAttention feeds the current query and the metadata keywords of candidate chunks to the model, allowing it to perform semantic matching and select the top-5 most relevant memory chunks for subsequent inference. The selected chunks are then incorporated into the model input, effectively augmenting the prompt with relevant long-term memory. This retrieval mechanism is inspired by the relevance-oriented text indexing design in Claude Code.\\
\indent A key design choice in MemAttention is to use the same model-driven semantic matching mechanism for both cross-chunk maintenance and inference-time retrieval. During compaction, the model matches the new chunk against the metadata of previously stored chunks to identify related memories for joint update, refinement, or stale-memory removal. During inference, it matches the current query against candidate chunk metadata and selects the top-$k$ relevant chunks to load, providing a unified notion of relevance across memory maintenance and retrieval.\\
\begin{algorithm}[t]
\small
\caption{MemAttention: Incremental Chunk Compaction and Model-Driven Retrieval}
\label{alg:memattention}
\begin{algorithmic}[1]
\State \textbf{Input:} new turn $x_t$, active chunk $C_a$, chunk table $\mathcal{T}[u][s]$, gate $\tau_c$, top-$p$ ($p{=}5$)
\State Append $x_t$ to $C_a$ \Comment{accumulate fresh context}
\If{$|C_a| \ge \tau_c$}
    \State $m \gets \textsc{Compress}(C_a)$ \Comment{compact one bounded chunk}
    \State $K_m \gets \textsc{BuildMeta}(m)$ \Comment{extract keywords / representative metadata}
    \State $\mathcal{M} \gets \{(\textit{cid}_i, K_i) \mid c_i \in \mathcal{T}[u][s]\}$
    \State $R \gets \textsc{LLMSelect}(K_m, \mathcal{M}, p)$ \Comment{model selects top-$p$ related prior chunks}
    \State $(m', U, D) \gets \textsc{JointCompact}(m, R)$ \Comment{refine new chunk; update/delete stale memory}
    \State Apply updates $U$ and deletions $D$ to $\mathcal{T}[u][s]$
    \State $K_{m'} \gets \textsc{BuildMeta}(m')$
    \State Insert $(u, s, \textit{cid}, m', K_{m'}, t_{\mathrm{now}})$ into $\mathcal{T}[u][s]$
    \State $C_a \gets \emptyset$ \Comment{start a new active chunk}
\EndIf
\State $q \gets \textsc{CurrentContext}(x_t)$
\State $K_q \gets \textsc{BuildQuery}(q)$
\State $\mathcal{M} \gets \{(\textit{cid}_i, K_i) \mid c_i \in \mathcal{T}[u][s]\}$
\State $S \gets \textsc{LLMSelect}(K_q, \mathcal{M}, p)$ \Comment{model selects top-$p$ relevant memory chunks}
\State $S \gets \textsc{FetchByID}(S, \mathcal{T}[u][s])$
\State $S \gets \textsc{SortByTime}(S)$ \Comment{restore chronological order}
\State \Return $\textsc{Concat}(S, q)$
\end{algorithmic}
\end{algorithm}
\heading{Algorithm overview.} Algorithm~\ref{alg:memattention} summarizes the workflow of MemAttention. Here, $u$ and $s$ denote the current \texttt{user\_id} and \texttt{session\_id}, respectively, and $\mathcal{T}[u][s]$ denotes the chunk table scoped to that user-session namespace. The active chunk $C_a$ accumulates newly arrived turns, $\tau_c$ is the chunk-compaction gate, and $p$ denotes the number of chunks selected by the model (default $p{=}5$). \emph{Lines~1--13} implement chunk production and cross-chunk reconciliation. Each new turn $x_t$ is first appended to the active chunk. Compaction is triggered only when the chunk length reaches the gate threshold $\tau_c$, at which point the system compresses the bounded chunk into a memory representation and extracts its metadata keywords. MemAttention then gathers the metadata of previously compacted chunks in the same namespace $\mathcal{T}[u][s]$ and feeds them, together with the metadata of the new chunk, to the model for semantic matching. The model selects the top-$p$ related prior chunks for \emph{joint compaction}, which allows the system to refine the new chunk and to update or delete stale prior memories when the new chunk contains corrections, refinements, or superseding information. The reconciled chunk is then written back to the chunk table, and a new active chunk is started. \emph{Lines~14--20} implement inference-time retrieval. The current context is converted into query keywords, and the system again collects candidate chunk metadata from the current namespace $\mathcal{T}[u][s]$. Instead of using an explicit scoring function, MemAttention feeds the query keywords and candidate metadata to the model, which performs semantic matching and selects the top-$p$ most relevant memory chunks. The selected chunks are fetched by their identifiers, reordered chronologically, and concatenated with the current context to form the final model input. By reusing the same model-driven semantic matching mechanism for both compaction-time reconciliation and inference-time retrieval, MemAttention maintains a unified notion of memory relevance while keeping both maintenance and retrieval overhead bounded.\\
\subsection{Memory Manager}
\label{sec:block-space-manager}
\begin{figure}[t]
    \centering \includegraphics[width=.98\columnwidth,height=0.2\textheight]{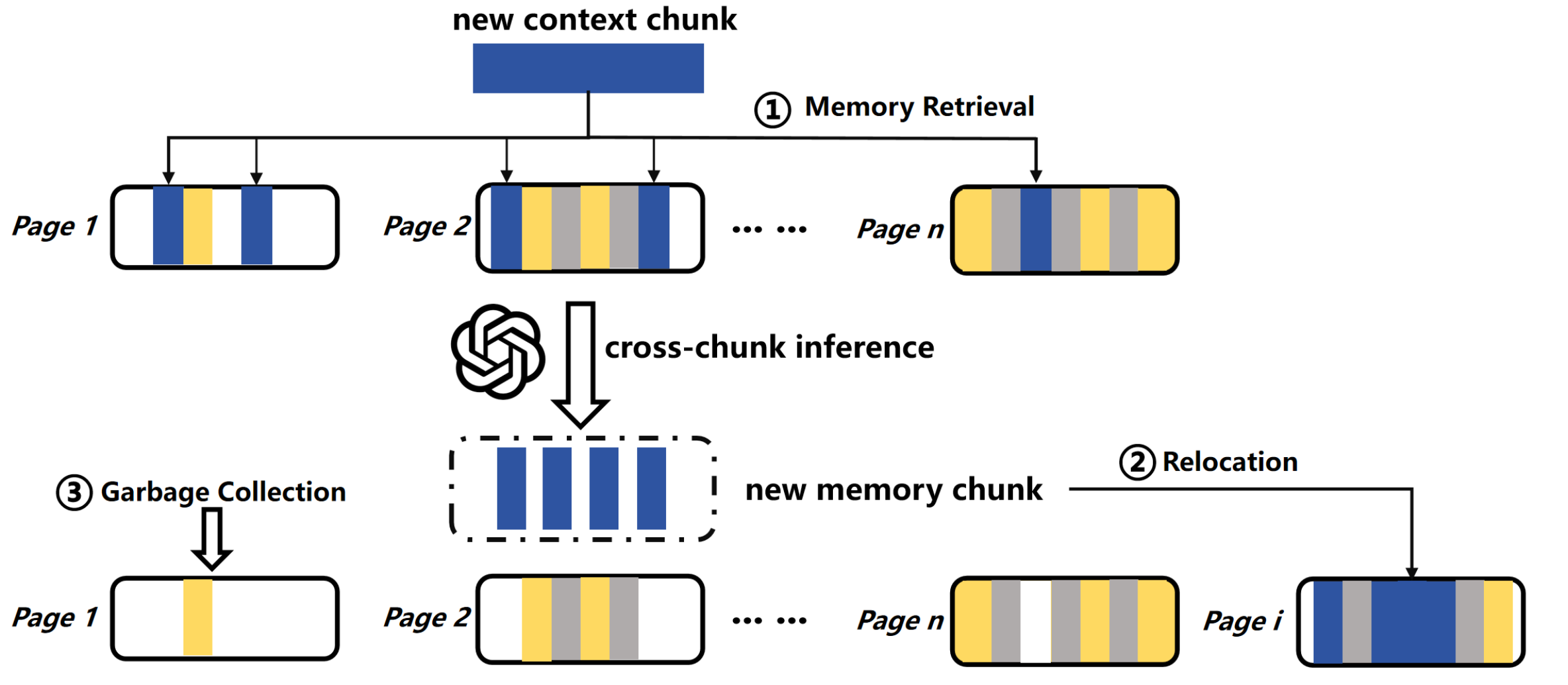}
    \caption{Overview of the Memory Compaction workflow.}
    \label{fig:memory_compaction}
    \vspace{-10pt}
\end{figure}
Although MemAttention is efficient at inference, its end-to-end performance still critically depends on how memory chunks are physically organized and maintained in storage. MemAttention stores memory as chunk records in a disk-resident vector store. Each chunk is immutable once written and is identified logically by its
\texttt{user\_id}, \texttt{session\_id}, and \texttt{metadata}, whereas its
physical location is determined by a block-level layout maintained by the
Memory Manager. This \textbf{separation between logical identity and physical placement
is crucial}~(\S\ref{sec:locality-gap-agent-memory}): it allows the system to reorganize chunk layout online without
changing the higher-level memory abstraction exposed to retrieval and inference.\\
\indent The goal of the Memory Manager is to reduce \emph{read amplification during
cross-chunk inference}. A cross-chunk inference step often needs to load multiple related
chunks together, but these chunks may have been written at different times and
thus scattered across different physical blocks. As a result, a single
inference may trigger multiple block reads even when the retrieved memories are
strongly related semantically and temporally. To address this inefficiency,
MemAttention performs \emph{association-aware relocation}: whenever joint
inference reveals that a set of chunks is frequently used together or has high
pairwise association, the system rewrites these chunks into a new block so that
future accesses can be served with fewer physical reads.
To drive this reorganization, MemAttention does not rely on a hand-crafted association score. Instead, it uses the model to infer chunk affinity directly from metadata, following the relevance-oriented retrieval strategy used in Claude Code~(\S\ref{sec:MemAttention}). The inferred affinity is then used to determine whether multiple chunks should be placed in the same physical block, thereby improving locality for future cross-chunk inference. Figure~\ref{fig:memory_compaction}
illustrates this workflow: \ding{182}~the system first retrieves related chunks
for a new context chunk and performs cross-chunk inference to produce a new
memory chunk; \ding{183}~it then relocates chunks that should be co-located
into new block(s); and \ding{184}~it finally reclaims invalidated space through garbage collection.

MemAttention adopts an out-of-place update policy. Once the model identifies a
set of chunks that should be co-located, the Memory Manager writes them into
new block(s) and marks their old copies as invalid, rather than rewriting
blocks in place. This design follows the same high-level principle as
LSM-tree compaction: updates are accumulated through sequential writes, stale
entries are invalidated through logical tombstoning, and background compaction
later reclaims space by preserving only live records~\cite{oneil1996lsm,chang2006bigtable}.
Let $\phi(B)$ denote the invalid ratio of block $B$:
\begin{equation}
\label{eq:invalid_ratio}
\phi(B) = \frac{N_{\mathrm{invalid}}(B)}{N_{\mathrm{total}}(B)} \, .
\end{equation}
When $\phi(B) \ge 0.75$, i.e., when invalid chunks occupy more than 75\% of a
block's capacity, the block becomes a candidate for compaction.

The compaction process serves two purposes. First, it frees space occupied by
obsolete chunk copies. Second, it further improves future locality by
repacking the remaining valid chunks according to both namespace and
model-inferred affinity. Specifically, the Memory Manager scans the candidate
block, extracts all valid chunks, partitions them by
$(\texttt{user\_id}, \texttt{session\_id})$, and then invokes the model again
within each partition to group chunks that are likely to be consumed together.
Each group is then packed into one or more new blocks while preserving
temporal order. This design preserves the logical isolation of different users
and sessions while still exploiting affinity within each namespace. As a
result, compaction is not merely garbage collection; it is also a layout
optimization pass that consolidates live, related memory into fewer blocks.\\
\heading{Memory Manager algorithm.}
Algorithm~\ref{alg:memory_manager} summarizes the Memory Manager. The algorithm
has two tightly coupled stages. The first stage runs on the joint-inference
path: it uses the model to identify retrieved chunks that should be physically
co-located, rewrites them out of place into new block(s), updates the logical
directory, and marks the old versions invalid. The second stage runs in the
background: when the invalid ratio of a block reaches 75\%, the block is
compacted by extracting its live chunks, grouping them by user and session,
and repacking them into new blocks according to model-inferred affinity.
~\emph{Lines~3--8} implement the online relocation path. Given the chunk set
$S$ retrieved for a joint inference, the Memory Manager invokes
$\textsc{LLMCoLocate}(S)$, which uses chunk metadata to identify the subset of
retrieved chunks that is likely to be co-accessed again. These chunks are then
packed into one or more new blocks out of place, their old physical copies are
marked invalid, and the logical directory is updated to point future reads to
the new locations. This step transforms co-access patterns observed at
inference time into improved physical locality for subsequent requests. \emph{Lines~11--24} implement background block compaction. For each block whose invalid
ratio satisfies $\phi(B) \ge 0.75$, the Memory Manager extracts only the live
chunks and partitions them by \texttt{user\_id} and \texttt{session\_id}. This
ensures that compaction never mixes chunks across tenants or sessions. Within
each partition, the system invokes $\textsc{LLMGroup}(P)$ to group chunks with
high model-inferred affinity, and repacks each group into one or more new
blocks while preserving temporal order. After the live chunks have been
rewritten and the directory updated, the original block is released. This
process both reclaims space and continuously reshapes the on-disk layout to
match evolving joint-inference access patterns.
\begin{algorithm}[t]
\small
\caption{Model-Driven Relocation and Block Compaction}
\label{alg:memory_manager}
\begin{algorithmic}[1]
\State \textbf{Input:} retrieved chunks $S$, block directory $\mathcal{D}$
\State \textbf{Output:} updated block layout
\State $G \gets \textsc{LLMCoLocate}(S)$ \Comment{model identifies chunks worth co-locating}
\If{$G \neq \emptyset$}
    \State $\mathcal{R} \gets \textsc{PackIntoNewBlocks}(\textsc{SortByTime}(G))$ \Comment{may span multiple blocks}
    \ForAll{$(c, B') \in \mathcal{R}$}
        \State $\textsc{MarkInvalid}(\mathcal{D}[c])$ \Comment{tombstone old copy}
        \State $\textsc{UpdateDir}(\mathcal{D}, c, B')$
    \EndFor
\EndIf
\ForAll{block $B$ with $\phi(B) \ge 0.75$}
    \State $V \gets \textsc{LiveChunks}(B)$ \Comment{extract valid memory only}
    \State $\mathcal{P} \gets \textsc{PartitionByUserSession}(V)$
    \ForAll{partition $P \in \mathcal{P}$}
        \State $\mathcal{G} \gets \textsc{LLMGroup}(P)$ \Comment{group chunks by model-inferred affinity}
        \ForAll{group $g \in \mathcal{G}$}
            \State $\mathcal{R}' \gets \textsc{PackIntoNewBlocks}(\textsc{SortByTime}(g))$
            \ForAll{$(c, B') \in \mathcal{R}'$}
                \State $\textsc{UpdateDir}(\mathcal{D}, c, B')$
            \EndFor
        \EndFor
    \EndFor
    \State \textsc{Release}(B) \Comment{old block is fully reclaimed}
\EndFor
\end{algorithmic}
\end{algorithm}
\subsection{Implementation}
\label{sec:impl}
 We implement \sysname by extending the control path of vLLM
v0.10.0~\cite{vllm_v0100_github}, while leaving the GPU
execution path, the attention backend, and the KV-cache
manager unchanged. The implementation follows a simple
separation of concerns. Memory retrieval is inserted at the
request admission layer, memory write-back is inserted at the
request completion layer, and physical layout maintenance is
split into an online relocation path and a background storage
manager. This design keeps long-term memory management
outside the hot token-level GPU decoding loop and allows
\sysname to be realized with limited changes to the serving
stack.

At the request admission layer, \sysname augments each
incoming request with a memory-scoped context before
tokenization. This context carries the stable namespace of the
request, including \texttt{user\_id}, \texttt{session\_id}, and
optionally \texttt{turn\_id}, together with request-level
controls such as \texttt{enable\_retrieval},
\texttt{enable\_writeback}, and
\texttt{memory\_budget\_tokens}. Given this context, the
memory retriever searches compacted chunks within the
corresponding \texttt{(user\_id, session\_id)} namespace,
collects their metadata, and feeds the current query together
with candidate metadata to the model for semantic matching.
The model returns the identifiers of the top-$p$ most
relevant chunks (default $p{=}5$); the system then fetches
these chunks, restores chronological order, and prepends
them to the prompt under the memory token budget. After
this step, the request follows the original vLLM execution
path without any change to scheduling or decoding.

At the request completion layer, \sysname intercepts finished
requests and performs incremental memory maintenance. Each
completed turn is appended to the active chunk of its
namespace. When the accumulated length exceeds
\texttt{chunk\_gate\_tokens}, the system triggers
chunk-gated compaction, rewrites the active chunk into a
compacted memory chunk, and builds its metadata. The
system then feeds this metadata together with the metadata of
previously compacted chunks in the same namespace to the
model, which selects the top-$p$ prior chunks for
reconciliation. Joint compaction then refines the new chunk
and updates or removes stale prior memories when the new
chunk contains corrections, refinements, or superseding
information. The reconciled result is written back to the
chunk store, while obsolete or invalidated chunks are marked
stale.

\sysname further introduces a Memory Manager to optimize
physical locality. During joint inference, the system records
which chunks are materialized together. Using the metadata
of jointly accessed chunks, the storage manager invokes the
model to identify subsets that should be co-located and
relocates them out of place into fresh block(s), while
invalidating their old copies. Separately, a background
compaction thread scans blocks whose invalid ratio satisfies
$\phi(B) \ge 0.75$, extracts live chunks, partitions them by
\texttt{(user\_id, session\_id)}, invokes the model within
each partition to group chunks likely to be consumed
together, repacks each group into new block(s) while
preserving temporal order, and reclaims obsolete blocks. In
this way, \sysname separates logical memory construction from
physical memory placement: the former is driven by
session-scoped semantic maintenance, while the latter is
driven by model-inferred co-access affinity and space
reclamation.

Overall, \sysname realizes long-horizon memory with modest changes to the vLLM control path while leaving the underlying decoding engine unchanged. The core implementation parameters are the chunk compaction gate (\texttt{chunk\_gate\_tokens}), the model selection width (\texttt{top\_p}=5), the retrieval memory budget (\texttt{memory\_budget\_tokens}), and the block reclamation threshold (\texttt{gc\_invalid\_ratio}=0.75).

\begin{figure*}[t]
    \centering
    \includegraphics[width=0.98\textwidth,height=6cm]{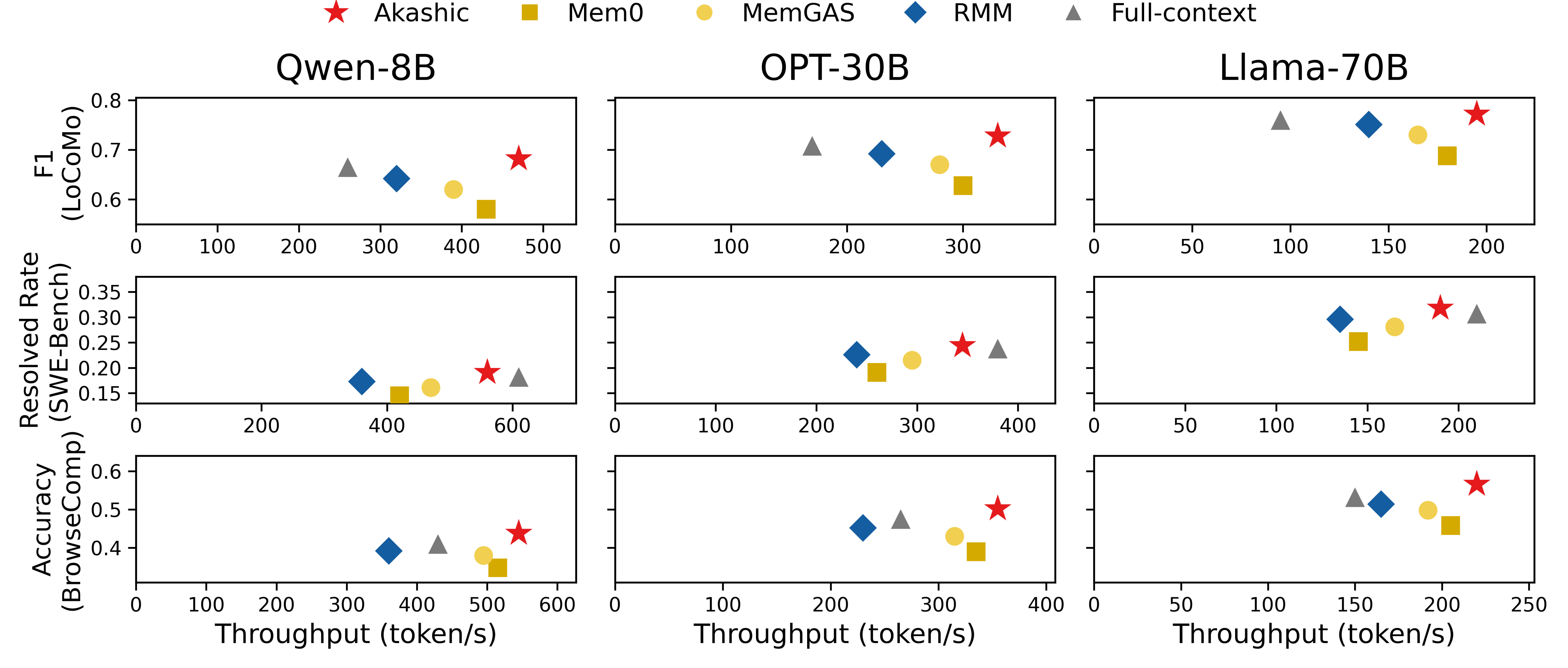}
    \vspace{-8pt}
    \caption{Throughput--accuracy trade-off under single-request inference (batch size $=1$) across three models and three workloads.}
    \vspace{-10pt}
    \label{fig:single-sequence}
\end{figure*}
\section{Evaluation}
\label{sec:eval}
In this section, we evaluate the performance of \sysname across a diverse set of workloads and models.
\subsection{Experimental Setup}
\label{subsec:exp-setup}
\begin{table}[t]
\centering
\small
\setlength{\tabcolsep}{3pt}
\caption{Models and server configurations.}
\label{tab:hardware_config}
\begin{tabular}{lccc}
\toprule
 & Qwen-8B & OPT-30B & Llama-70B \\
\midrule
GPUs & 1$\times$H800 & 1$\times$H800 & 2$\times$H800 \\
GPU mem. & 80\,GB & 80\,GB & 160\,GB \\
Weights & 16\,GB & 61\,GB & 140\,GB \\
Free mem. & 64\,GB & 20\,GB & 20\,GB \\
SSD & 4\,TB NVMe & 4\,TB NVMe & 4\,TB NVMe \\
PCIe link & Gen5$\times$16 & Gen5$\times$16 & 2$\times$Gen5$\times$16 \\
PCIe bw. & $\sim$64\,GB/s & $\sim$64\,GB/s & $\sim$128\,GB/s \\
\bottomrule
\end{tabular}
\vspace{-10pt}
\end{table}

\heading{Model and server configurations.}
We use Qwen~\cite{yang2025qwen3}, OPT~\cite{zhang2022opt}, and Llama~\cite{grattafiori2024llama3} models with 8B, 30B, and 70B parameters, respectively, for our evaluation. These three models cover small, medium, and large open-weight LLM configurations and allow us to study the behavior of \sysname across substantially different memory footprints. We deploy the 8B and 30B models on a single NVIDIA H800 GPU, and the 70B model on two NVIDIA H800 GPUs. All servers are equipped with PCIe-attached H800 accelerators and local NVMe SSD storage and we use OPT-30B to perform relevance analysis over memory chunks. The detailed model sizes and hardware configurations are shown in Table~\ref{tab:hardware_config}.

\heading{Workloads.}
We evaluate \sysname on four representative long-horizon workloads:
LoCoMo~\cite{maharana2024locomo} (long-term multi-session dialogue memory),
SWE-bench~\cite{jimenez2024swebench} (high-density software engineering trajectories),
BrowseComp~\cite{wei2025browsecomp} (bursty web browsing trajectories),
and WebArena~\cite{zhou2024webarena} (realistic multi-step web-agent tasks).
Together, these datasets cover the main workload regimes discussed in our motivation~(\S\ref{sec:3.1}).

\heading{Baselines.} We compare \sysname against five baselines:
Full-context (no compression or external memory),
Mem0~\cite{mem0} (whole-context memory summarization),
MemGAS~\cite{xu2025memgas} (segment-level multi-granularity memory),
MemGPT~\cite{packer2023memgpt} (hierarchical external memory),
and RMM~\cite{rmm} (reflective memory refinement).
Together, these baselines span the whole-context, segment-level, and hierarchical memory strategies discussed in \S\ref{sec:motivation}.

\heading{Key metrics.}
We use accuracy and throughput as the two primary evaluation metrics. Accuracy is measured using each benchmark's canonical protocol: QA F1~\cite{maharana2024locomo}, \%Resolved~\cite{jimenez2024swebench}, Accuracy~\cite{wei2025browsecomp}, and task success rate~\cite{zhou2024webarena}.

\subsection{Overall Effectiveness}
\label{sec:eval:ThroughputandAccuracyImprovements}
We evaluate \sysname with basic sampling on the models and workloads above. Figure~\ref{fig:single-sequence} shows the throughput--accuracy trade-off under basic sampling (\ie, batch size $=1$). \sysname consistently lies on the \textbf{Pareto frontier} across all nine settings. On LoCoMo and BrowseComp, \sysname improves throughput over Mem0, MemGAS, and RMM by 1.08$\times$--1.10$\times$, 1.10$\times$--1.21$\times$, and 1.33$\times$--1.54$\times$, while improving accuracy by 8.4--10.2, 4.2--7.2, and 2.1--5.2 points, respectively. Compared with full-context inference, \sysname is 1.27$\times$--2.05$\times$ faster and still improves accuracy by 1.2--3.4 points. \emph{The reason is} that \sysname summarizes memory at chunk granularity, so each maintenance step processes only a bounded unit instead of the entire history, which reduces memory-construction overhead. At the same time, its cross-chunk retrieval and joint reconciliation recover semantically related evidence that would otherwise be split across segments, allowing the final prompt to remain short while still containing the critical information needed for inference.\\
\indent The same trend holds on SWE-Bench, although the throughput gap to full-context inference is smaller. \sysname improves throughput over Mem0, MemGAS, and RMM by 1.31$\times$--1.33$\times$, 1.15$\times$--1.19$\times$, and 1.41$\times$--1.56$\times$, while improving resolved rate by 4.6--6.6, 2.9--3.7, and 1.8--2.2 points, respectively. Compared with full-context inference, \sysname retains 90.5\%--91.8\% of the throughput while still improving resolved rate by 0.6--1.1 points. This smaller throughput gap is expected because SWE-Bench is more information-dense and therefore less compressible. Even in this regime, \sysname remains preferable because Mem0 compresses the whole history too aggressively and mixes irrelevant content, MemGAS keeps each segment short but loses cross-segment dependencies, and RMM preserves stronger semantics at the cost of heavier memory-maintenance overhead. In contrast, \sysname keeps memory compact enough for efficient serving while using cross-chunk recovery to preserve the information that most directly affects task accuracy.
\subsection{Robustness under Heterogeneous Density and Concurrency}
\label{sec:eval:Robustness}
\begin{figure*}[t]
    \centering
    \includegraphics[width=0.98\textwidth,height=6cm]{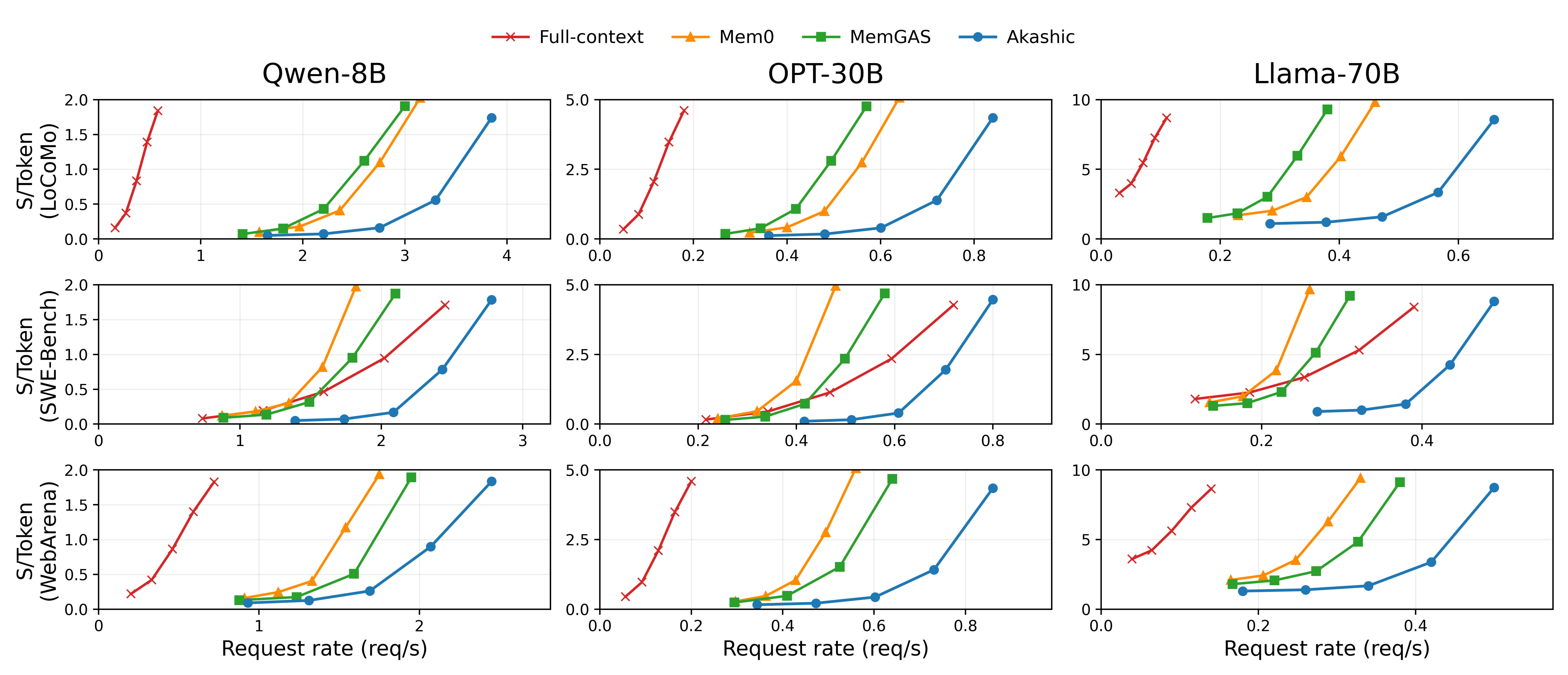}
    \vspace{-8pt}
    \caption{\sysname maintains stable performance as concurrency increases, across models of different scales and diverse workloads.}
    \vspace{-10pt}
    \label{fig:RobustnessofAkachic}
\end{figure*}
We next evaluate \sysname under concurrent serving by increasing the offered request rate and measuring average latency per generated token (s/token). As shown in Figure~\ref{fig:RobustnessofAkachic}, \sysname consistently shifts the knee of the latency curve furthest to the right across all three models and all three workloads, indicating the highest sustainable request rate before latency rises sharply. The gains are largest on LoCoMo: \sysname sustains 1.22$\times$--1.43$\times$ higher request rates than Mem0, 1.28$\times$--1.74$\times$ higher than MemGAS, and 4.67$\times$--6.64$\times$ higher than Full-context. These improvements come from bounded chunk-level maintenance, which rewrites only the active chunk instead of repeatedly summarizing the full history, together with memory compression and cross-chunk recovery, which retain relevant evidence without lengthening the prompt and thus avoid the large prefill cost of long retained contexts. \\
\indent The same trend appears on WebArena, where \sysname delivers 1.40$\times$--1.54$\times$ gains over Mem0, 1.26$\times$--1.34$\times$ over MemGAS, and 3.40$\times$--4.30$\times$ over Full-context. These gains remain substantial because WebArena trajectories are long, bursty, and semantically heterogeneous, so short prompts and efficient retrieval reduce both prompt-side and retrieval-side overhead under load. On SWE-Bench, the margin is smaller but still consistent: \sysname sustains 1.53$\times$--1.88$\times$ higher request rates than Mem0, 1.32$\times$--1.58$\times$ higher than MemGAS, and 1.11$\times$--1.26$\times$ higher than Full-context. This smaller gap is expected because SWE-Bench traces are more information-dense and therefore less compressible. Even so, \sysname still degrades more gracefully by keeping memory maintenance bounded and retrieving only task-relevant evidence rather than carrying unnecessary context. Overall, Figure~\ref{fig:RobustnessofAkachic} shows that efficient memory construction and retrieval, rather than full-history retention, are key to sustaining high-throughput LLM serving under concurrency.
\subsection{Locality and Storage Efficiency}
\begin{figure}[t]
    \centering \includegraphics[width=.98\columnwidth,height=0.22\textheight]{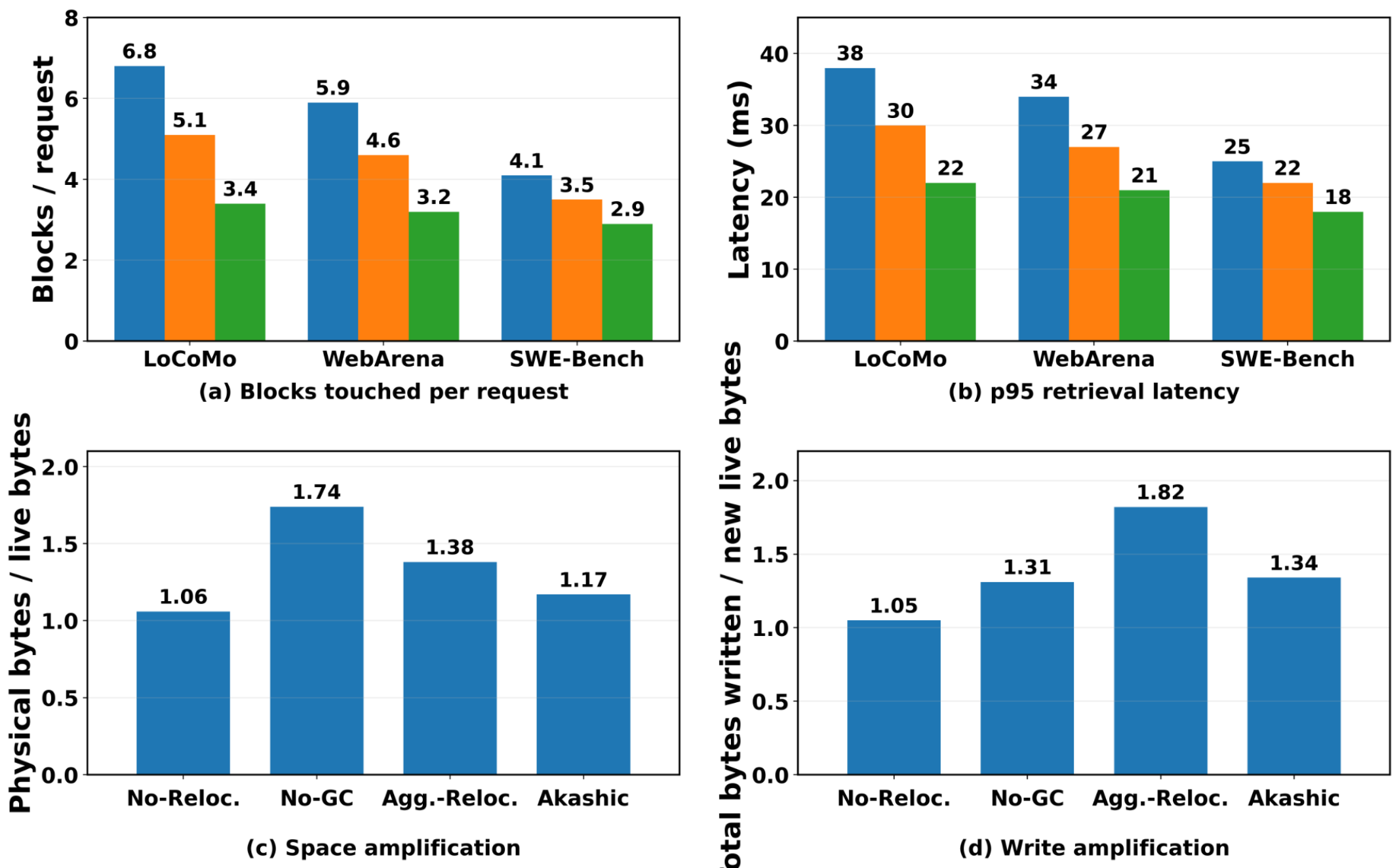}
    \caption{Compared with append-only and semantic-only layouts, \sysname reduces the number of blocks touched per request and lowers p95 cold-cache retrieval latency by co-locating jointly accessed chunks. Meanwhile, selective out-of-place relocation with background garbage collection keeps both space amplification and write amplification moderate, showing that improved locality does not come at excessive storage overhead.}
    \vspace{-18pt}
    \label{fig:Localityefficiency}
\end{figure}
This experiment asks whether \sysname's Memory Manager can close the locality gap in \S\ref{sec:locality-gap-agent-memory} without excessive storage overhead. We replay the same inference traces with the same chunking policy, query-aware retrieval logic, and memory budget, varying only the physical layout policy. Figures~\ref{fig:Localityefficiency}(a)--(b) compare \textsc{Append-only}, \textsc{Semantic-only}, and \textsc{\sysname}; Figures~\ref{fig:Localityefficiency}(c)--(d) compare \textsc{No-Relocation}, \textsc{No-GC}, \textsc{Aggressive-Relocation}, and \textsc{\sysname} under the same update stream.

Figure~\ref{fig:Localityefficiency}(a) shows the main locality benefit: by colocating chunks that are jointly accessed at inference time, \sysname reduces blocks/request on LoCoMo, WebArena, and SWE-Bench from 6.8/5.1 to 3.4, 5.9/4.6 to 3.2, and 4.1/3.5 to 2.9 (\textsc{Append-only}/\textsc{Semantic-only} $\rightarrow$ \textsc{\sysname}), corresponding to reductions of 50.0\%/33.3\%, 45.8\%/30.4\%, and 29.3\%/17.1\%. Figure~\ref{fig:Localityefficiency}(b) shows the latency effect of this improved locality: p95 cold-cache retrieval latency falls from 38/30 to 22\,ms, 34/27 to 21\,ms, and 25/22 to 18\,ms, i.e., by 42.1\%/26.7\%, 38.2\%/22.2\%, and 28.0\%/18.2\% relative to \textsc{Append-only}/\textsc{Sema}\\
\textsc{ntic-only}. The gains are largest on LoCoMo and WebArena, whose longer and more fragmented histories make co-access-aware placement more effective than write-order placement or semantic-only grouping.

Figures~\ref{fig:Localityefficiency}(c)--(d) show that these locality gains remain storage-efficient. \sysname achieves 1.17$\times$ space amplification and 1.34$\times$ write amplification, compared with 1.06$\times$/1.05$\times$ for \textsc{No-Relocation}, 1.74$\times$/1.31$\times$ for \textsc{No-GC}, and 1.38$\times$/1.82$\times$ for \textsc{Aggressive-Relocation}. Thus, \sysname is only 10.4\% above \textsc{No-Relocation} in space amplification, but 32.8\% below \textsc{No-GC} and 15.2\% below \textsc{Aggressive-Relocation}; for write amplification, it is 27.6\% above \textsc{No-Relocation}, only 2.3\% above \textsc{No-GC}, and 26.4\% below \textsc{Aggressive-Relocation}.\\ 
\indent Overall, \sysname closes the locality gap by physically colocating jointly accessed chunks and reclaiming obsolete copies in the background, reducing the cost of cross-chunk inference while maintaining a favorable balance between locality and storage overhead.
\subsection{Ablation and Sensitivity}
\label{Sec:5.5}
\begin{figure}[t]
    \centering \includegraphics[width=.98\columnwidth,height=0.3\textheight]{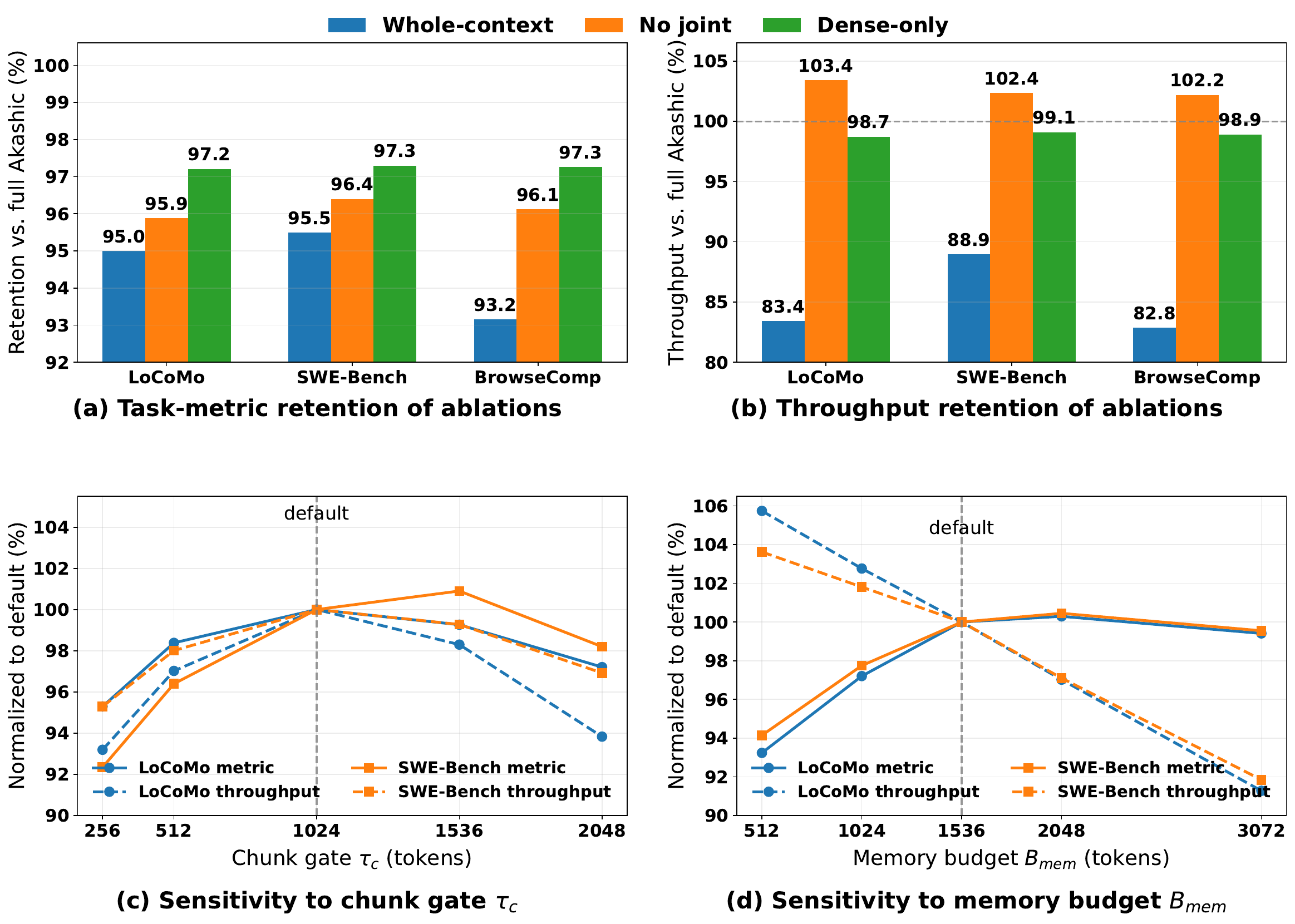}
    \caption{Ablation and sensitivity of \sysname.
(a) Task-metric retention under component ablations, normalized to full \sysname.
(b) Throughput retention under the same ablations, normalized to full \sysname.
(c) Sensitivity of task quality and throughput to the chunk gate $\tau_c$.
(d) Sensitivity of task quality and throughput to the memory budget $B_{mem}$.}
    \vspace{-10pt}
    \label{fig:Ablation_1}
\end{figure}
We next ablate \sysname's logical and physical components to identify which mechanisms are essential and how sensitive the system is to parameter choice. We study three logical components---bounded chunk-level maintenance, cross-chunk reconciliation, and model-driven relevance matching using OPT-30B---and sweep four key parameters: the chunk gate $\tau_c$, the memory budget $B_{mem}$, the association threshold $\tau_{assoc}$, and the GC invalid-ratio threshold $\theta_{gc}$. Figure~\ref{fig:Ablation_1}(a)--(b) reports task-quality and throughput retention, both normalized to full \sysname, and shows that bounded chunk-level maintenance is the dominant efficiency contributor. Replacing it with \textit{Whole-context} reduces task-quality retention to 95.0\%, 95.5\%, and 93.2\% on LoCoMo, SWE-Bench, and BrowseComp, while lowering throughput more sharply to 83.4\%, 88.9\%, and 82.8\%, respectively. This result indicates that \sysname's main efficiency gain comes from bounding each maintenance step to one chunk rather than repeatedly rewriting the full history. Removing cross-chunk reconciliation (\textit{No joint}) preserves slightly higher throughput (102.2\%--103.4\%) but reduces task quality to 95.9\%--96.4\%, indicating that reconciliation mainly improves evidence recovery at modest overhead. \textit{Dense-only} remains closer to the full system in both quality (97.2\%--97.3\%) and throughput (98.7\%--99.1\%), suggesting that simple embedding-based similarity already captures a substantial fraction of useful relevance signals, whereas \sysname's model-driven relevance matching further improves selection quality by identifying semantically related chunks more accurately during both reconciliation and retrieval. Figure~\ref{fig:Ablation_1}(c)--(d) further shows that \sysname is not brittle to parameter choice. Both workloads peak near the default $\tau_c{=}1024$: smaller chunks fragment evidence and trigger compaction more often, whereas larger chunks weaken semantic isolation and increase per-compaction cost. Increasing $B_{mem}$ from 512 to 1536 improves quality from 93.2\% to 100\% on LoCoMo and from 94.1\% to 100\% on SWE-Bench, but the gains saturate beyond the default, while throughput falls to 91.3\% and 91.8\% at 3072.\\
\begin{figure}[t]
    \centering \includegraphics[width=.98\columnwidth,height=0.15\textheight]{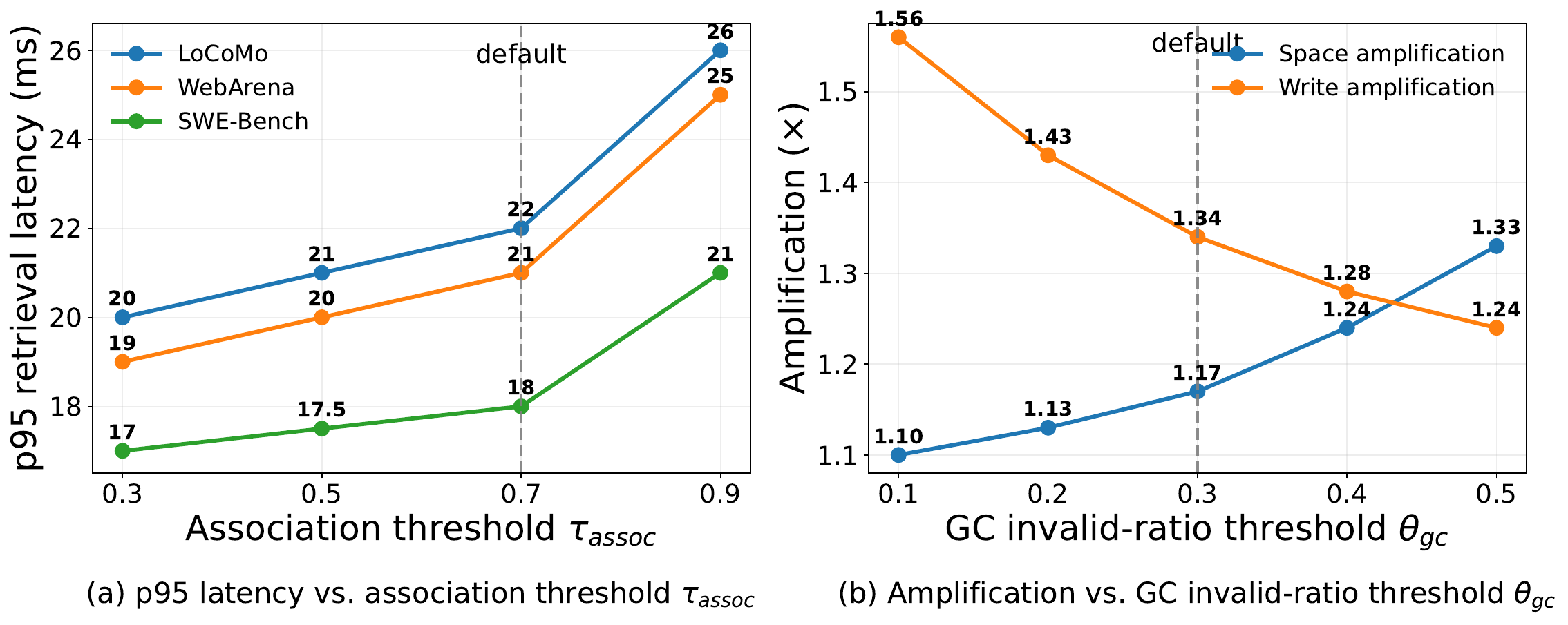}
    \vspace{-5pt}
    \caption{Sensitivity of \sysname's storage manager.
(a) Sensitivity of p95 retrieval latency to the association threshold $\tau_{assoc}$.
(b) Sensitivity of space and write amplification to the GC invalid-ratio threshold $\theta_{gc}$.}
    \vspace{-15pt}
    \label{fig:Ablation_2}
\end{figure}
\indent Figure~\ref{fig:Ablation_2} then examines the sensitivity of \sysname's physical storage manager. Figure~\ref{fig:Ablation_2}(a) shows that lowering $\tau_{assoc}$ makes relocation more aggressive and slightly reduces p95 retrieval latency, while the default $\tau_{assoc}{=}0.7$ remains close to the latency knee at 22/21/18\,ms on LoCoMo/WebArena/SWE-Bench; increasing $\tau_{assoc}$ further to 0.9 degrades latency to 26/25/21\,ms. Figure~\ref{fig:Ablation_2}(b) shows the expected GC trade-off: as $\theta_{gc}$ increases from 0.1 to 0.5, write amplification drops from 1.56$\times$ to 1.24$\times$, but space amplification rises from 1.10$\times$ to 1.33$\times$. The default $\theta_{gc}{=}0.3$ therefore provides a balanced operating point at 1.17$\times$ space amplification and 1.34$\times$ write amplification. Overall, \sysname's gains do not come from a single knob. Bounded chunk maintenance delivers most of the efficiency benefit, reconciliation improves quality at modest overhead, and locality-aware storage preserves these gains while maintaining a balanced trade-off among retrieval latency, space amplification, and write amplification.

\section{Related Work}
\label{sec:related_work}
\heading{Agent Applications and Workloads.} LLM agents have shown utility across tool-augmented reasoning, embodied exploration, software engineering, web navigation, and long-horizon dialogue~\cite{wang2023llm_agents_survey,yao2023react,wang2023voyager}. Representative benchmarks include SWE-bench for resolving real GitHub issues~\cite{jimenez2024swebench}, WebArena and BrowseComp for realistic web interaction and persistent search~\cite{zhou2024webarena,wei2025browsecomp}, and LoCoMo for temporal and causal reasoning over long multi-session dialogues~\cite{maharana2024locomo}. Although these workloads differ in form, they impose a common systems challenge: long-horizon agents continuously accumulate context that must be filtered, retrieved, and served efficiently.\\
\heading{Memory Systems for Long-Horizon Agents.} Prior work largely treats long-term agent memory as a problem of semantic construction and retrieval. Recursive summarization compresses history into progressively shorter summaries~\cite{recursive}, whereas MemGPT introduces an OS-inspired hierarchy that pages between in-context and external memory~\cite{packer2023memgpt}. Recent systems further explore persistent memory extraction and retrieval, including Mem0~\cite{mem0}, SeCom~\cite{pan2025secom}, RMM~\cite{rmm}, MemGAS~\cite{xu2025memgas}, and A-MEM~\cite{xu2025amem}. \sysname complements this line of work by jointly optimizing logical memory construction and physical layout: maintenance is bounded at chunk granularity, reconciliation and retrieval share the same relevance function, and co-accessed chunks are physically colocated to reduce retrieval cost.\\
\heading{LLM Serving and Locality-Aware Storage.} LLM serving engines such as Orca and vLLM optimize batching, scheduling, and GPU and KV-cache management for high-throughput decoding~\cite{yu2022orca,kwon2023pagedattention}, but they do not treat persistent agent memory as a first-class serving object. At the storage layer, disk-resident ANN and vector-database systems such as SPANN, FreshDiskANN, and Starling study I/O efficiency, freshness, and layout locality under large-scale search workloads~\cite{chen2021spann,singh2021freshdiskann,wang2024starling}. \sysname adopts this systems perspective at a different boundary: it manages dynamically evolving, session-scoped memory chunks whose semantic relevance and physical layout must be co-optimized. In this sense, \sysname connects agent-memory systems with locality-aware storage for long-horizon LLM serving.
\section{Conclusion}
This paper has presented \sysname, a low-overhead LLM inference service for long-horizon agents. To the best of our knowledge, \sysname is the first system to efficiently realize long-context filtering for LLM agents through end-to-end software--hardware co-design. Across LoCoMo, SWE-bench, BrowseComp, and WebArena, \sysname consistently stays on the Pareto frontier against prior memory baselines, improving task accuracy by 8.4--10.2 points, increasing throughput by 1.21$\times$--1.35$\times$ under basic sampling, and sustaining 1.26$\times$--1.74$\times$ higher load under concurrent serving. These gains come from jointly optimizing chunk-granular memory construction, cross-chunk reconciliation, query-aware retrieval, and locality-aware storage management. Upon acceptance, we plan to open-source the code, artifacts, and reproduction materials.

\bibliographystyle{ACM-Reference-Format}
\bibliography{reference}
\end{document}